\documentclass[journal]{IEEEtran}

\usepackage{graphicx}
\usepackage{color}
\usepackage{amsmath}
\usepackage{bbm}
\usepackage{hyperref}

\begin{document}

\title{Learning the image processing pipeline}

\author{Haomiao Jiang, Qiyuan Tian, Joyce Farrell, Brian Wandell \\
Department of Electrical Engineering, Stanford University \\
Psychology Department, Stanford University}

\maketitle

\begin{abstract}
Many creative ideas are being proposed for image sensor designs, and these may be useful in applications ranging from consumer photography to computer vision. To understand and evaluate each new design, we must create a corresponding image processing pipeline that transforms the sensor data into a form that is appropriate for the application. The need to design and optimize these pipelines is time-consuming and costly. We explain a method that combines machine learning and image systems simulation that automates the pipeline design. The approach is based on a new way of thinking of the image processing pipeline as a large collection of local linear filters. We illustrate how the method has been used to design pipelines for novel sensor architectures in consumer photography applications.
\end{abstract}

\begin{IEEEkeywords}
Local Linear Learned, Camera image processing pipeline, machine learning
\end{IEEEkeywords}

\IEEEpeerreviewmaketitle

\section{Introduction}
\IEEEPARstart{A}{dvances} in small, low-power CMOS image sensors and related optics have revolutionized consumer photography\cite{Gamal02, Beatriz16, Fossum14, Chang12, Reshidko16}. These technologies have improved dramatically the spatial resolution, dynamic range, and low light sensitivity of digital photography.

In addition to improving conventional photography, these technologies open up many possibilities for novel image systems architectures. The new optics and CMOS sensors capabilities have already motivated novel camera architectures that extend the original Bayer RGB design. For example, in recent years a new generation of architectures have been produced to increase spatial resolution\cite{Longoni08}, control depth of field through light field camera designs\cite{Georgiev12, Bishop12, Marwah13}, extend dynamic range and sensitivity by the use of novel arrangements of color filters\cite{Baranov15} and mixed pixel architectures \cite{Nayar15, Yasuma10}.

To develop these opportunities requires that we innovate on the third fundamental component of image systems, the image processing pipeline. The pipeline is the set of algorithms, including demosaicking, noise reduction, color management, and display nonlinear transforms (gamma curves), that convert the sensor data into a rendered image. Even modest changes to the camera architecture, such as more color pixels into the mosaic \cite{Monno12} or including near infrared detectors \cite{Tang15} can require substantial rethinking of the image processing pipeline. New image processing pipelines, specialized for the new types of cameras, are slow to develop. Consequently, the design of new imaging sensors is far outpacing the development of algorithms that can take advantage of these new designs\cite{Nayar06}, and the vast majority of image processing algorithms are still designed for sensors that use the classic single plane Bayer RGB spatial sampling mosaic.  

In this paper, we describe a new framework that enables image systems engineers to rapidly design image processing pipelines that are optimized for novel camera architectures. The general idea of the framework was first proposed by Lansel et al. in 2011\cite{Lansel11}. Here, we introduce the framework in the form of a set of software tools that use simulation and learning methods to design and optimize image processing pipelines for these new camera systems.

This paper is organized into three sections that define our contributions. First, we explain the image processing architecture: the input data are grouped by their local features into one of a set of local classes, where locality refers to both position on the sensor array (space), pixel type (color) and response level. The optimal affine transform in each class is learned using camera simulation technology. We refer to this framework as $L^3$ to emphasize its key principles: Local, Linear and Learned. Second, we assess the performance of $L^3$ method by comparing the rendered quality with the ones from high-end modern digital cameras. We specifically show that such a collection of affine transforms accurately approximates the complex, nonlinear pipelines implemented in modern consumer photography systems. Third, we illustrate how $L^3$ method can learn image processing pipelines for new camera architectures.

There are a number of related efforts that incorporate system joint optimization and data-driven learning methods in designing camera image processing pipeline. We discuss the relationship between $L^3$ and these contributions more fully in Discussion section after introducing $L^3$ framework.

\section{The \texorpdfstring{$L^3$}{L3} Method}
The $L^3$ method comprises two main steps: rendering and learning. The rendering step adaptively selects from a stored table of affine transformations to convert the raw camera sensor data into an image in the target color space (e.g. sRGB). The training step learns and stores the transformations used in rendering.

Conventional image processing pipelines often include nonlinear elements, including thresholding operations and 'gamma' transforms\cite{Poynton93, Guo04}. The $L^3$ rendering algorithm uses a collection of affine transforms that are applied to data in the different classes. The accuracy of a collection of affine transforms for approximating the image processing pipeline, including nonlinearities such as the 'gamma', can be controlled by the number of classes; this is the conventional local linear approximation to continuous functions. The challenge of managing discrete transitions, such as the transition from the linear response region to saturation, represented by thresholds can be managed by selecting proper category boundaries. As a practical matter, there is rarely a strong need to specify a precise and sharp category boundary in natural image processing.

In the following, we explain these two steps of $L^3$ method in detail. We explain the method using the example of a camera sensor with RGBW (red, green, blue and white) color filter array. The method can be applied to other designs, which we describe in the Results section.

\subsection{Rendering}
\begin{figure}[!htbp]
\begin{center}
\includegraphics[width=0.48\textwidth]{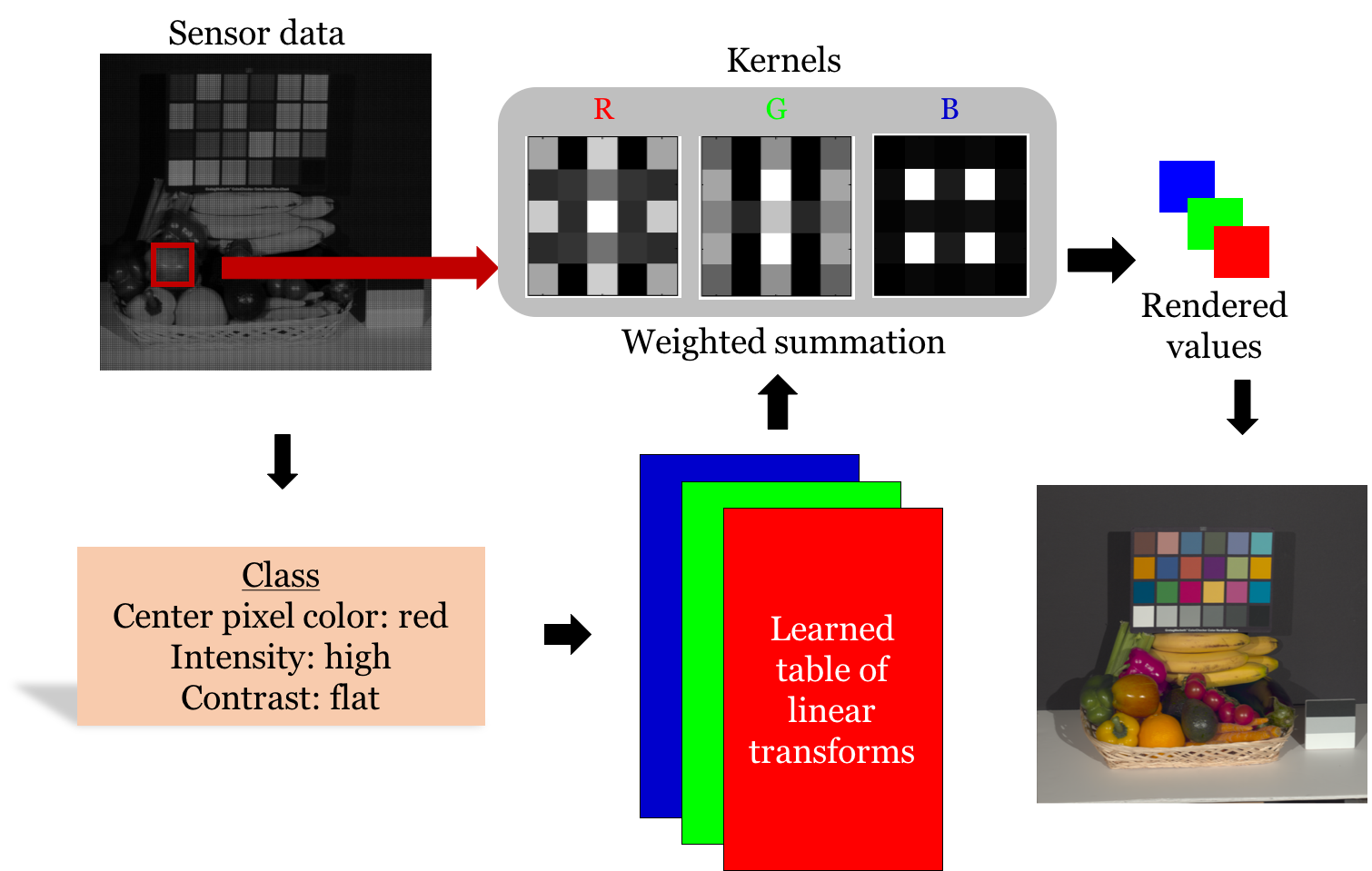}
\end{center}
\caption{\textbf{The $L^3$ rendering method.} In $L^3$ rendering, each pixel is classified by its local features into one of a few hundred classes (e.g. a red pixel, with high intensity, surrounded by a uniform field). Each class has a learned affine transform (stored in a table) that weights the sensor pixel values in order to calculate the rendered outputs. Hence the rendered output of each pixel is calculated as an affine transform of the pixel and its neighbors.}
\label{Fig:L3Render}
\end{figure}

The rendering pipeline begins with the sensor data in the spatial neighborhood of a pixel, $n(x, y, p)$. We illustrate the method for a $5\times5$ neighborhood, so that the neighborhood comprises 25 pixel responses (Figure \ref{Fig:L3Render}). Each pixel is classified into one of many classes, c, based on its local features: the identity of the pixel, the mean response level of the local patch, the spatial variance of the neighborhood, and pixel saturation status, etc. The total number of classes can be estimated as the product of the number of categories for every feature. For example, suppose that there are 4 types of pixels, and we categorize the mean neighborhood intensity into 10 response levels. This produces 40 different classes. If we further classify each neighborhood into textured (high variance) or uniform, then there will be 80 classes ($1\leq c\leq 80$). Additionally, pixel saturation is also a condition that requires careful management. In some designs, one type of pixels (e.g., the $W$ pixel in an RGBW) can saturate while the RGB pixels will be well within their operating range. It is important to create separate classes that account for pixel saturation. Allowing for this increases the number of effective classes, typically by a factor of three for the RGBW case. The number of classes is a hyperparameter of the $L^3$ framework and can vary in different applications.

For each class $c$ and output channel $r$, we retrieve an affine transform to map the 25 values into one rendered value. The affine transforms for each class and rendered output channel, $T(c, r)$, are pre-computed and stored. The rendered output, $o(x, y, r)$ is the inner product of the stored affine transform and the neighborhood data (augmented with a 1 to account for the affine term).

$$o(x,y,r)= <T(c,r), n(x,y,p)>$$

There are several practical considerations in $L^3$ implementation: the content and application dependent class definitions, the target color representations (monochrome, highly saturated) and even the fitting model in each class (affine or polynomial). These choices impact the algorithm efficiency and precision, and we describe experiments evaluating different choices in Results. These choices are part of the design process when implementing the $L^3$ method.

Finally, the computations for each pixel are independent, meaning that the architecture can be highly parallelized with a graphical processing unit (GPU) or other hardware to shorten rendering time\cite{Tian15}. And $L^3$ is designed to perform real-time rendering for high quality images on mobile devices.

\subsection{Learning}
It is challenging to create algorithms for cameras that are being designed, rather than cameras that already exist, because of limitations in obtaining sensor data\cite{Khashabi14}. We solve this problem by using image systems simulation tools to model the proposed camera architecture and to create the training data\cite{Farrell04, Farrell12}. The Image Systems Engineering Toolbox (ISET) begins with a spectral representation of the scene and includes simulations of the optics and sensor. The simulator has been validated against data from several real devices \cite{Farrell08,Chen09}.

Once we have simulations of the critical data, we have a variety of options for how we select the transforms that map the sensor data into the rendered images. We describe our approach to simulation and transformation derivation in the next two sections.

\subsubsection{Training data}
Realistic camera training requires scene spectral radiance data sets that are representative of the likely application. We have obtained scene spectral radiance and accumulated examples from a number of public scene radiance data sets\footnote{http://www.imageval.com/scene-database/}. In addition, we have also used spectral computer graphics methods to simulate a variety of scene spectral radiance images. These simulations can produce scene spectral radiance examples for training that establish special spatial resolution and color requirements that extend what we would be likely to find by merely sampling a range of natural images. 

\begin{figure}[!htbp]
\begin{center}
\includegraphics[width=0.48\textwidth]{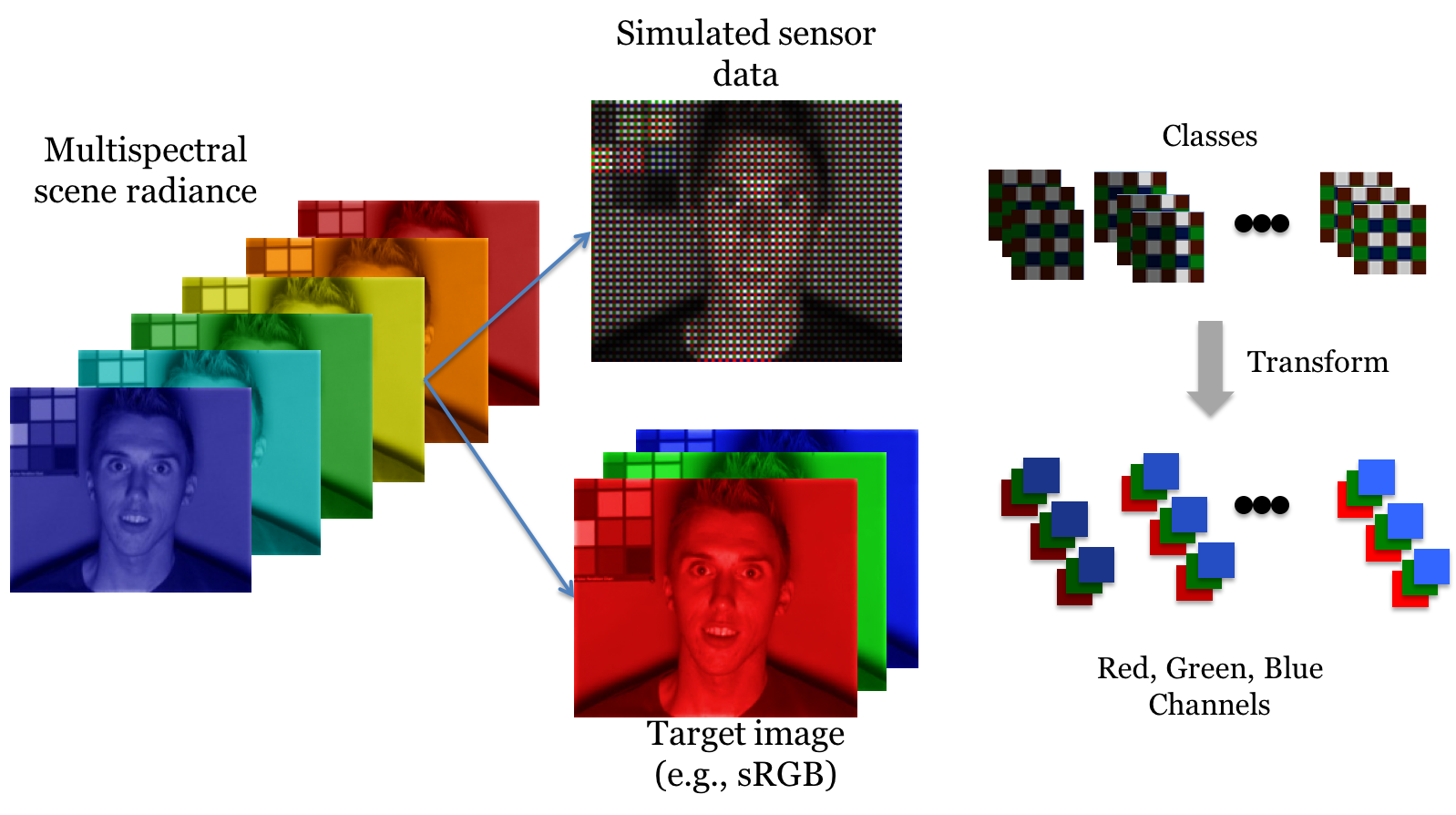}
\end{center}
\caption{\textbf{Preparing training data with image systems simulation.}  Starting from scene spectral radiance data (left), we compute two aligned representations. Top: the sensor responses in a model camera system. Bottom: the target rendered image from scene radiance. The simulated sensor data at each pixel and its neighborhood (patch) is placed into one of several hundred pre-defined classes, based on properties such as the pixel type, response level, and local image contrast. For each class, the pairs of patch data and target output values are used to learn an affine transformation from the data and the target output value.
}
\label{Fig:L3Train}
\end{figure}

A further advantage of the simulation method is that the desired output image and sensor data are precisely aligned, at the pixel level. For each input scene spectral radiance we calculate the calibrated color representation (e.g. CIE XYZ) and the sensor response at each pixel (Figure \ref{Fig:L3Train}). Such correspondence is very difficult or even impossible to obtain from empirical measurements. 

Finally, because the training pairs are produced by simulation, we can produce many examples of scenes and measurement conditions. Through simulation we can control the properties of the ambient lighting, including its level and spectral power distribution. We can perform simulations with a wide range of optical parameters. Training can be performed for special content (e.g., faces, text, or outdoors scenes).

The pairs of images produced by the simulation methods can provide a virtually limitless collection of input data to the training system.

\subsubsection{Transform optimization}
The purpose of the training is to calculate a set of transforms that optimally map the sensor responses to the target output image and minimize the empirical risk. Stated formally, our task is to find for each class, $C_i$, a transformation, $T_i$ such that
$$\underset{T_i}{\text{minimize}} \sum_{j\in C_i} \mathcal{L}(y_j, X_jT_i) $$

Here, $X_j$ is a row vector containing the $j$-th patch data from the sensor; $y_j$ are the corresponding target rendered image values. $\mathcal{L}$ is the loss function (error) between the target image and the transformed sensor data. In consumer imaging applications, the visual difference measure CIE $\Delta E_{ab}$\cite{Luo01} can be a good choice for the loss function. For other applications, regularized RMSE is widely used. In nearly all photography applications, the transformation from sensor to rendered data is globally non-linear. $L^3$ method approximates the global nonlinear transformation with a collection of affine transforms for appropriately defined classes $C_i$.

At first, we solve the transforms for each class independently. This problem can be expressed in the form of ordinary least-squares. To avoid noise magnification, we use ridge regression and regularize the kernel coefficients. That is

$$ T_i= \underset{T_i}{\text{argmin}} ||\tilde{y}-XT_i||_2^2+ \lambda ||T_i||_2^2$$

The data from each patch are placed in the rows of $X$; the regularization parameter is $\lambda$, and $\tilde{y}$ is the output in the target color space as an $N\times 3$ matrix. We have experimented using several target color spaces, including the XYZ, CIELAB and sRGB representations, and we can find satisfactory solutions in all cases. The closed-form solution for this problem is given as

$$ T_i=(X^TX + \lambda I)^{-1}X^T\tilde{y}$$

The computation of $T_i$ can be further optimized by using singular vector decomposition (SVD) of $X$. That is, if we decompose $X = UDV^T$, we have

$$T_i = V\text{diag}(\frac{D_j}{D_j^2+\lambda})U^T\tilde{y}$$

The regularization parameter ($\lambda$) is chosen to minimize the generalized cross-validation (GCV) error \cite{Golub79}.

Once the transforms for each class are defined, it is possible to review them for properties that reflect our prior knowledge, such as continuity over the sensor input space, symmetry and uniformity (see Discussion). The software implementation includes methods to check for these conditions and to bring the transforms into alignment with this knowledge.

\section{Results}
In this section, we characterize the performance of the $L^3$ method and illustrate how it can be used to generate image processing pipelines for novel camera architectures. First, we analyze whether the $L^3$ rendering method based on many affine transforms is sufficient to approximate the performance of commercial image processing pipelines (Nikon and DxO). Second, we use $L^3$ to learn a collection of transforms for non-standard color sensor designs (RGBW and RGB-NIR).

\subsection{The \texorpdfstring{$L^3$}{L3} pipeline closely approximates high quality Bayer CFA algorithms}
The $L^3$ pipeline is designed to be computationally efficient and to learn algorithms for novel arrays. Before applying the method to new designs, it is important to analyze whether the simple pipeline is capable of supporting high quality rendering expected from camera systems that have been optimized. To evaluate any performance limits, we compared how well the $L^3$ rendering algorithm can approximate the image processing pipeline embedded in a high quality commercial camera. 

\subsubsection{Accuracy}
In one experiment, we used an image dataset of 22 well-aligned raw and JPEG natural images from a Nikon D200 camera. We used 11 randomly selected images for training the local linear transforms and the other half for testing (cross-validation). The $L^3$ pipeline parameters were set to use 50 luminance levels for the four pixel types (red, blue and two types of green), for a total of 200 classes. We analyzed the data with $5\times5$ local patches (affine transforms of 26 parameters). The effect of patch size is discussed later. 

\begin{figure}[!htbp]
\begin{center}
\includegraphics[width=0.48\textwidth]{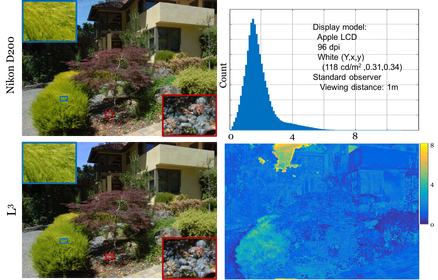}
\end{center}
\caption{\textbf{Approximating the Nikon pipeline with $L^3$ local linear transforms.} At the left, we compare camera RGB (upper left) with $L^3$ rendered image (lower left). The image at the lower right measures the perceptual difference between the two images using S-CIELAB $\Delta E_{ab}$ values for each pixel. The metric is calculated assuming that the images are displayed on a known, calibrated monitor (see inset text, upper right). The histogram of errors is shown on the upper right. The mean error is 1.84, the peak error is near 8, and the standard deviation of the $\Delta E_{ab}$ values is 0.9. The $L^3$ transforms were learned from one set of 11 images. This image is from an independent data set of 11 images. The errors reported for this image are typical for all the images in the independent test set. }
\label{Fig:L3Nikon}
\end{figure}

Figure \ref{Fig:L3Nikon} (upper left) shows a typical example of an image produced by the Nikon processing pipeline and the corresponding image produced by the $L^3$ method (lower left). By visual inspection the images are very similar; the largest visual differences are the blue sky and the bush in the lower left. We use a perceptual space-color visual difference metric S-CIELAB \cite{Zhang97} to quantify the visual difference. Perceptual metrics require specifying the display and viewing distance, and for the S-CIELAB calculation we assumed the images are rendered and viewed on a calibrated LCD monitor with $96$ dpi at viewing distance of one meter. For the $2592\times 3872$ Nikon images, the horizontal field of view is $58.7$ deg. The $\Delta E_{ab}$ error image (lower right) and histogram (upper right) are typical of the test data: the mean S-CIELAB $\Delta E_{ab}$ is 1.74 for the 11 test images (PSNR 40.36), which is a small visual difference. 

These experiments show that the collection of $L^3$ transforms approximates the full commercial rendering produced by this Nikon D200 camera for this collection of outdoor images.

We also applied the $L^3$ method to learn the local linear transforms that approximates a commercial image processing pipeline (DxO). In this experiment, 26 raw and RGB image pairs are analyzed. The RGB images are generated with DxO Optics Pro using parameters tuned by a professional photographer, Dave Cardinal. This dataset includes multiple types of cameras and the content spans various natural scenes, human portraits, and scenic vistas.

\begin{figure}[!htbp]
\begin{center}
\includegraphics[width=0.48\textwidth]{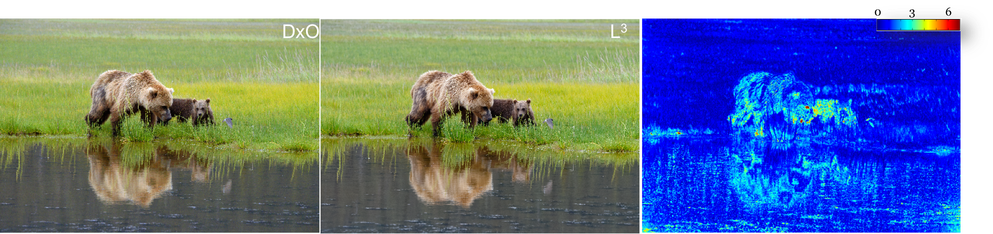}
\end{center}
\caption{\textbf{Approximating DxO pipeline with $L^3$ local linear transforms.} We compare the image rendered by a DxO pipeline with parameters tuned by a professional photographer (Left) with one rendered with the $L^3$ method (Middle). The perceptual error is measured in S-CIELAB $\Delta E_{ab}$ image (Right). The error calculations are based on the same monitor as in Figure \ref{Fig:L3Nikon}.}
\label{Fig:L3DxO}
\end{figure}

Each of the individual images can be well-approximated by the $L^3$ method. The images at the left and middle of Figure \ref{Fig:L3DxO} show a typical example of the DxO image and the $L^3$ image. The image at the right is the S-CIELAB visual difference error for each pixel. The mean S-CIELAB $\Delta E_{ab}$ value for this image is 1.458 and the accuracy of the approximation is similar to what we achieved for the Nikon processing pipeline. 

The expert’s settings vary significantly as the scene and camera types change; for example, in some scenes the expert chooses more sharpening and for others a softer focus is preferred. Hence, no single set of $L^3$ transforms applies to all of the images. The broad issue of selecting $L^3$ transforms for specific acquisition conditions or rendering aesthetics is further analyzed in Discussion. 

The DxO and Nikon D200 experiments show that the $L^3$ kernel regression approach is sufficient to approximate the transforms embedded in commercial rendering products.

\subsubsection{The transforms}
Next, we examine the properties of the learned transforms that approximate the Nikon D200 processing pipeline. When learning the $L^3$ transforms, a few parameters must be selected: the number and distribution of response levels, and the size of the local patch. 

\begin{figure}[!htbp]
\begin{center}
\includegraphics[width=0.48\textwidth]{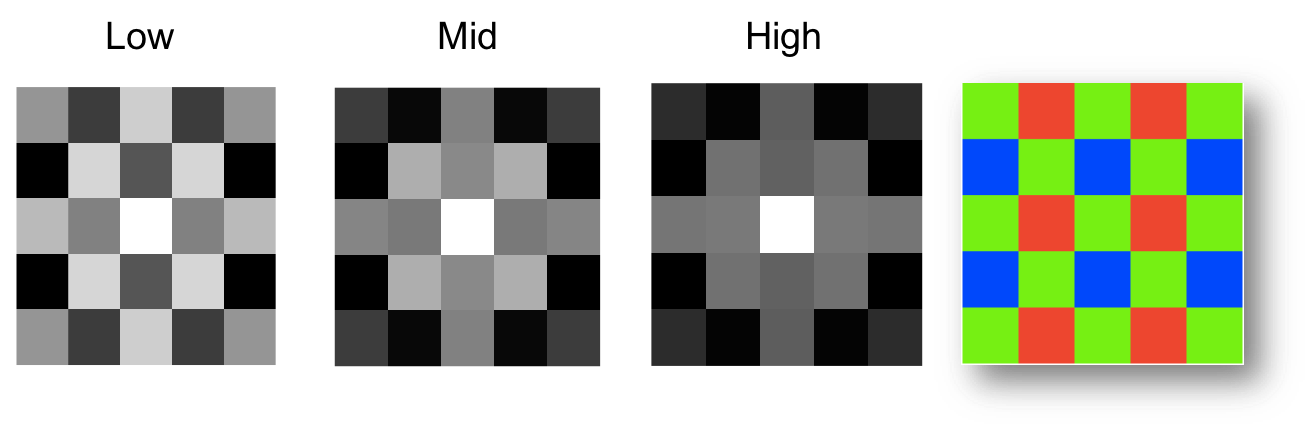}
\end{center}
\caption{\textbf{$L^3$ transforms depend on response level.} The three monochrome images show the relative weights of the transforms that convert data at a G pixel centered patch into the green output channel. The patch size is $5\times5$ and the three images show the weights that are learned for a class defined by Low (left), Mid (second left) and High (third left) response levels. The spatial distribution of the weights becomes more concentrated at the central pixel as the mean response level increases. The CFA pattern of the full $5\times5$ patch is shown at the right.}
\label{Fig:L3TransformBayer}
\end{figure}

The transforms change substantially with the response level (Figure \ref{Fig:L3TransformBayer}). At low levels, the weights are relatively equal across the entire patch, and there is less selectivity for color channels. At high response levels the weights are concentrated on the appropriate pixel type. For example, when the center pixel is green and the output channel is also green, at high response levels the transform weights are concentrated on the central green pixel. At low light levels the transform weights at non-central green pixels are relatively high.

\subsubsection{Response level spacing}
The learned transform weights change more rapidly at the lower response levels compared to the higher levels. For this reason, it is efficient to use a logarithmic spacing of the mean response levels that define the classes; that is, we use more finely spaced classes at the low response levels than the high response levels. 

\begin{figure}[!htbp]
\begin{center}
\includegraphics[width=0.48\textwidth]{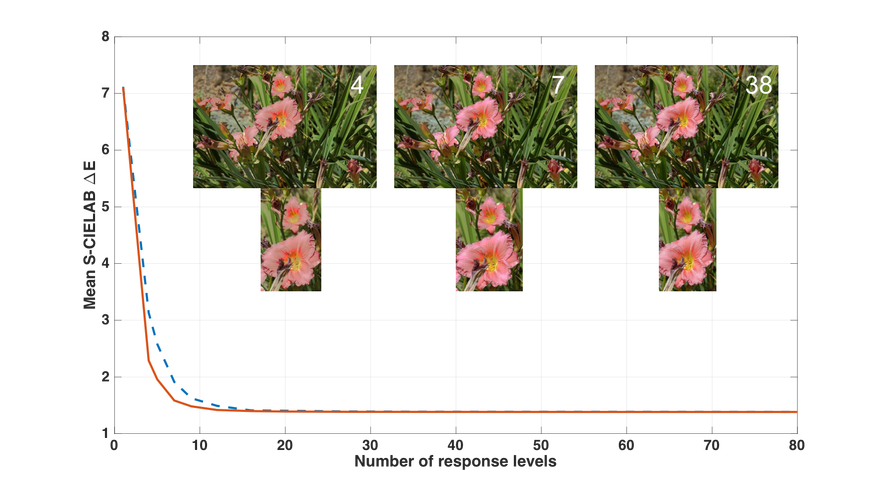}
\end{center}
\caption{\textbf{The effect of class selection on rendered image quality.} The graph shows how the color error (mean S-CIELAB $\Delta E$) of the rendered image declines as the number of response level classes increases. The two curves show the effect for linear (blue dashed) and logarithmic (red solid) spacing of the mean response levels in the patch. This inset images are the renderings for 4, 7 and 38 mean response level classes (logarithmic spacing). The rendering of the flower (zoomed view) changes substantially as the number of levels increases, and the mean color reproduction error declines significantly as the number of mean response levels increases to about 15. There is only a very small advantage for the logarithmic spacing when using a small number of classes, and no advantage beyond about 12 levels.}
\label{Fig:L3ClassesSpacing}
\end{figure}

Through simulation, we can evaluate the difference between linear and logarithmic spacing of the mean response levels. We analyzed the Nikon D200 data using different numbers of mean response levels (Figure \ref{Fig:L3ClassesSpacing}). The levels were spaced either linearly or logarithmically. To achieve the same image quality (e.g. 2 $\Delta E$), logarithmic spacing is equivalent to linear spacing with about 50\% more number of classes. As the total number of classes becomes large, say 50 for this example, the performance of the two spacing methods is very similar. We expect that the specific parameter values, such as number of response levels, will differ slightly for different optics and sensor combinations. But the principle of using logarithmic spacing is likely to hold across conditions.

\subsubsection{Patch size selection}
There is a significant computational cost to increasing patch size. Changing the patch size from a $5\times5$ to $7\times7$ ($9\times9$) approximately doubles (triples) the number of coefficients and computational cost. Moreover, if the training dataset is limited, the risk of overfitting can increase with the patch size. 

\begin{figure}[!htbp]
\begin{center}
\includegraphics[width=0.48\textwidth]{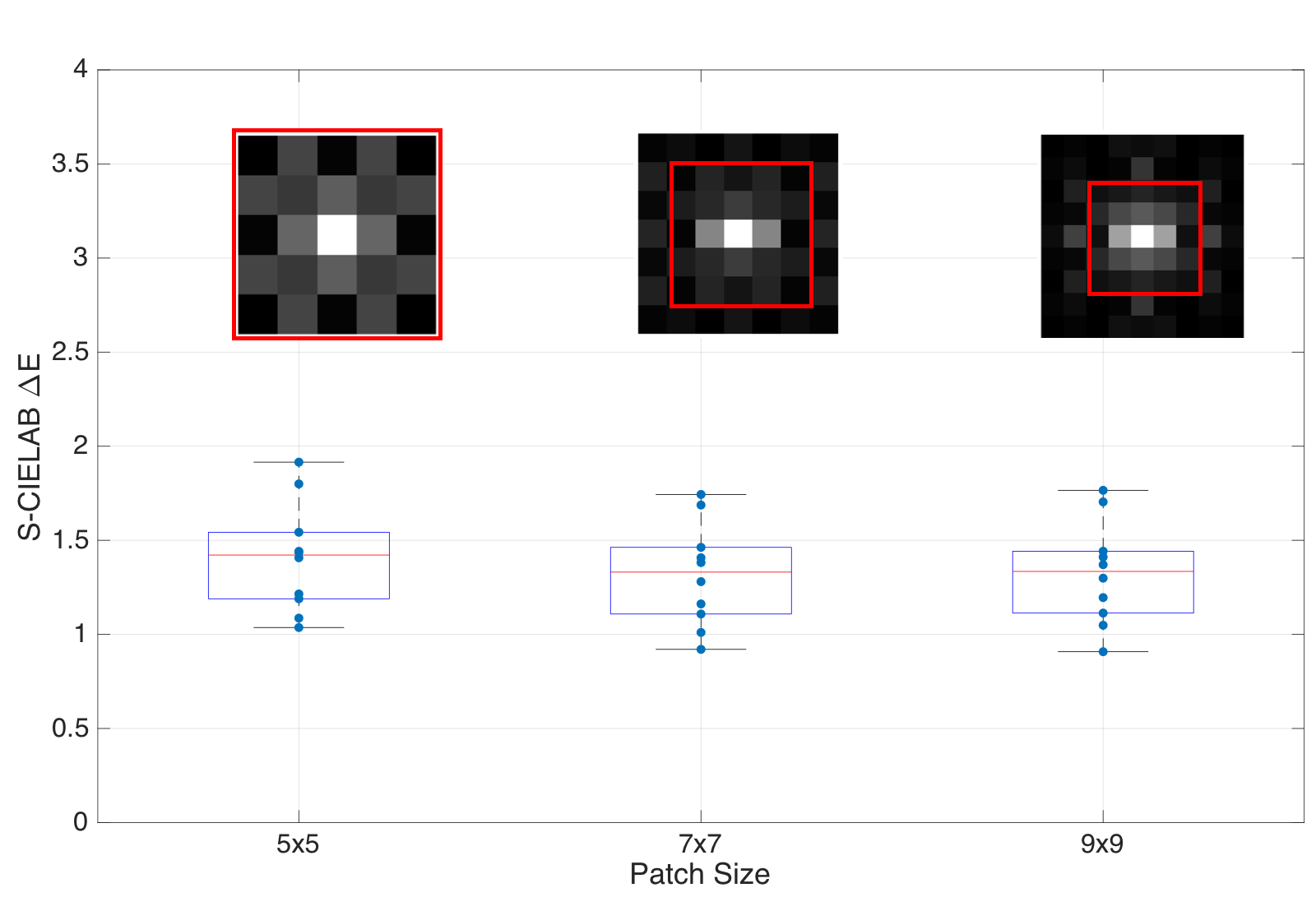}
\end{center}
\caption{\textbf{The effect of patch size on rendered image quality.} The boxplot shows the distribution of mean S-CIELAB $\Delta E$ errors for the 11 test images. The inset monochrome images are examples of the learned affine transforms. The red rectangle in each image denotes the center $5\times5$ region. The weights learned by the transform outside of the $5\times5$ patch are small, and increasing the patch size has little effect on the color accuracy.}
\label{Fig:L3PatchSize}
\end{figure}

For the Nikon and DxO approximations, we found little improvement in the approximation as we changed the patch size beyond $5\times5$. Figure \ref{Fig:L3PatchSize} shows the mean $\Delta E$ values of S-CIELAB on the 11 test images. The plotted data points show the mean error for individual test images, and the solid red line shows the average of all 11 images. The mean $\Delta E$ values decrease very slightly as the patch size increases from $5\times5$ to $7\times7$ and there is no further decrease as the patch size increases to $9\times9$. It might be more effective to create classes based on other features (e.g. nonlinear or global features) rather than increasing the patch size.

\subsection{\texorpdfstring{$L^3$}{L3} pipelines for novel color filter arrays}
The ability to automate the process of learning the rendering pipeline is an important objective of the $L^3$ method. In this section, we use $L^3$ to generate the image processing pipelines for two challenging CFA designs.

We first apply $L^3$ method to generate an image processing pipeline for camera with RGBW sensor. The CFA repeating pattern of RGBW sensor contains both RGB and clear (white) pixel. Adding a white pixel extends the operating range of the camera and makes the camera usable in low light imaging. The key challenge in designing a pipeline for this sensor is the large mismatch in the sensitivity between the W and RGB pixels\cite{Parmar09}

We then consider a CFA that combines RGB channels with a near infrared (NIR) channel. There is a great deal of interest in adding an NIR channel to support applications of depth sensing. The NIR channel, which is invisible to the human eye, can be used to measure a projected NIR pattern for depth estimation. The NIR pixels do not contain significant information for image reproduction, so that this design reduces the pixel count significantly. We analyze concerns how to best render an image for the RGB-NIR design, and how this rendering depends on factors such as pixel size and optics.

\subsubsection{RGBW}
\begin{figure}[!htbp]
\begin{center}
\includegraphics[width=0.48\textwidth]{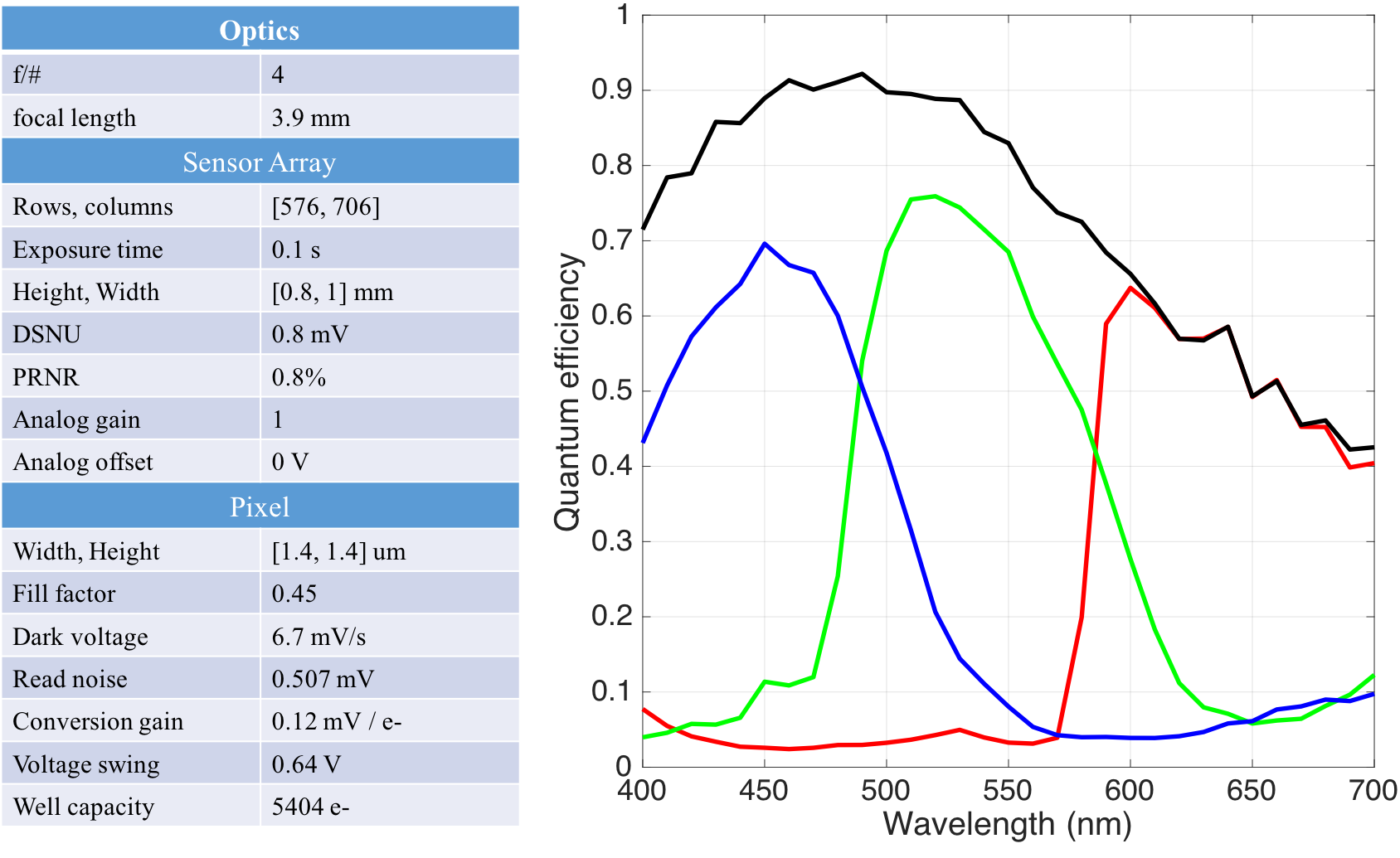}
\end{center}
\caption{\textbf{The simulated RGBW camera system.} The table lists key properties of the optics, pixel and sensor array used in the simulation. The curves show the spectral quantum efficiency of the four different pixel types. The inset shows the CFA pattern for a $5\times5$ patch centered on a green pixel. }
\label{Fig:L3RGBW}
\end{figure}

In this experiment, we simulated an RGBW camera with exactly one R, G, B and W in each CFA repeating pattern. The spectral transmittance of the color filters and other key sensor and lens parameters of the camera simulation are shown in Figure \ref{Fig:L3RGBW}. The relative sensitivity of the W to the RGB pixels and the spatial arrangement of the four pixel types differs between vendors \cite{Kumar09,Wang11}, and this simulation represents one of a range of possible choices.

We apply simulation methods (Figure \ref{Fig:L3Train}) to prepare the training data to learn the local linear transforms. Specifically, we first calculate the sensor response of multispectral scenes using the ISET camera simulator. Then, we compute the ideal (CIE XYZ) value at each pixel location. The local patches are classified by mean response levels and the center pixel type. Affine transformations are learned for each class, using ridge regression with the regularization parameters set using a cross-validation error minimization.

\begin{figure}[!htbp]
\begin{center}
\includegraphics[width=0.48\textwidth]{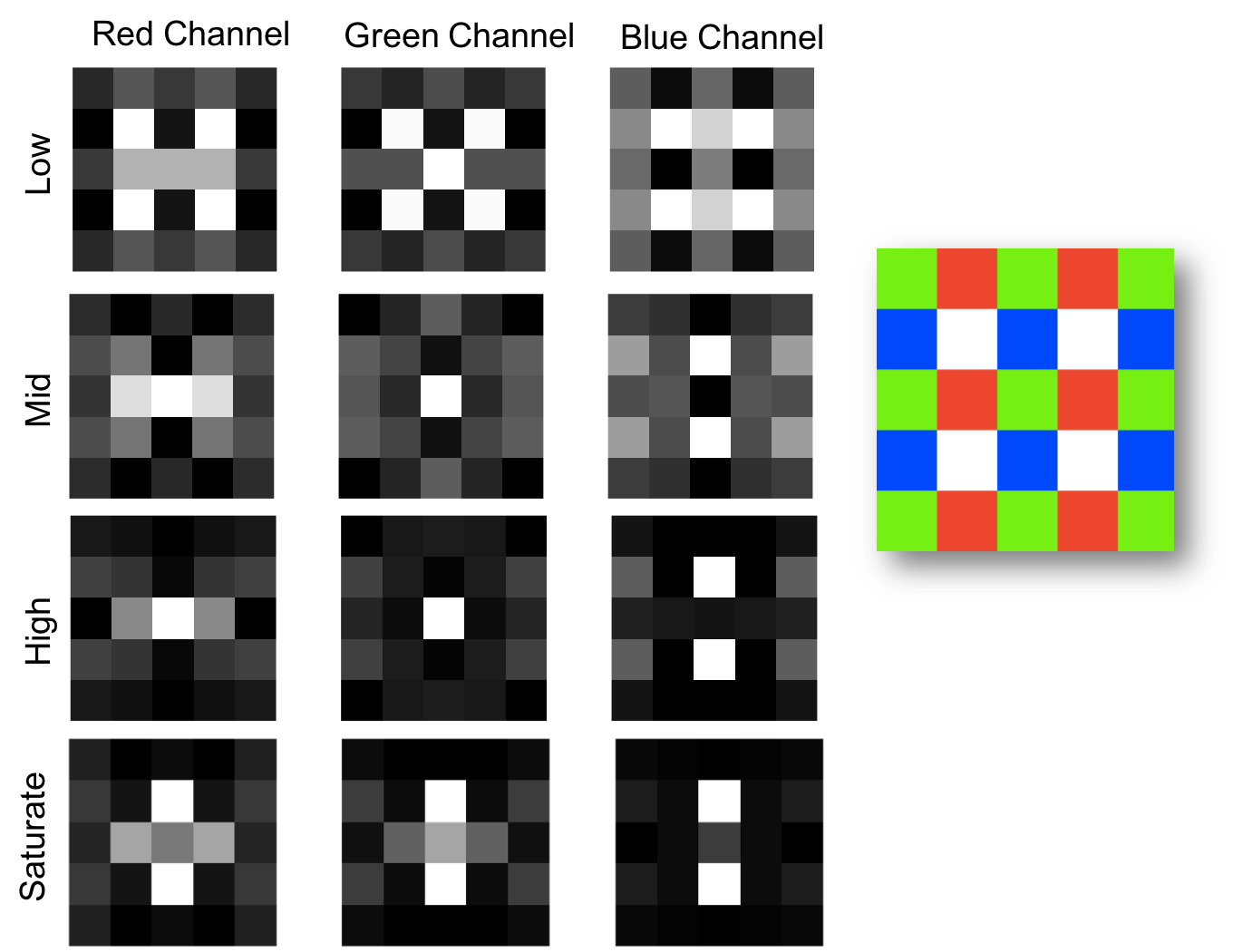}
\end{center}
\caption{\textbf{The transforms learned for the RGBW sensor.} The transforms learned for Low, Mid, High and Saturate response levels are shown in the rows. The columns show the red, green and blue output channels. The color filter array is shown at the right. At Low response levels, the highest weights are on the W pixels. As the response level increases, the weights of the RGB pixels become larger. When most W and some of the G pixels saturate, the R and B weights are the largest.}
\label{Fig:L3TransformRGBW}
\end{figure}

Figure \ref{Fig:L3TransformRGBW} shows examples of the learned transforms for this RGBW camera in four different classes, from low response levels to near saturation. The low mean response class transforms (Low) heavily weight data from the W pixel, presumably because the signal-to-noise ratio (SNR) of the W pixel is substantially higher at generally low response levels. As the response level increases (Mid), the W pixel SNR advantage is less important than the color information provided by the RGB pixels, and the weights redistribute to using more of the data from these pixels than the white pixel. As the W pixel saturates (High) the $L^3$ transforms further discounts the W responses. As the G pixel begins to saturate as well (Saturate), the weights on both the W and G pixels decline, with most of the weight being assigned to the R and B pixels. By designing tables that include many response levels, we assure a smooth transition from the W-dominated to the RGB-dominated domain.

Notice that for the Mid and High response levels, the red output depends significantly on the the G center pixel response. The algorithm learns that there is a strong spatial and color correlation between the G center pixel value and output red channel\cite{Hel-Or04}. This confirms the previous observation that the value at the G center pixel is useful in predicting the red output value, and the linear transform quantitatively estimates the proper amount that G pixel should contribute to the red channel output. However, as the center G pixel starts to saturate (Saturate), the transforms assign them lower weights.

We analyzed two important aspects of the $L^3$ performance: color accuracy and image resolution. We performed this analysis by training on a standard data set and then testing performance on targets designed to analyze color and resolution.

To assess color accuracy, we compute the CIELAB $\Delta E$ values between the $L^3$ rendered image and ideal values of a standard Macbeth color checker. The $L^3$ pipeline for the RGBW sensor achieves a mean $\Delta E$ of 1.7, which is very accurate. We then replaced the W pixel with a green pixel to form a traditional Bayer CFA pattern. The mean $\Delta E$ difference is similar ($\Delta E$= 1.65). Hence, the $L^3$ method learns how to incorporate the W pixels to achieve an accurate color reproduction. 

To assess resolution, we calculated the spatial frequency response (SFR) using the ISO 12233 slanted bar method\cite{ISO12233}. This method measures the image intensity in the region near the edge of a slanted bar. The intensity measurements are converted from a spatial representation into a modulation transfer function (MTF). The metric is defined by the spatial frequency at which the MTF drops to half of its peak value (MTF50); higher MTF50 values imply higher spatial resolution. The value of the MTF50 depends on a number of system features, including the optics and the pixel size. For the system described in Figure 8 (f/\#=4, pixel size 1.4 um), the MTF50 is 154.80 cycles/mm, which is close to the upper bound imposed by the optics 186.70 cycles/mm. 

\begin{figure}[!htbp]
\begin{center}
\includegraphics[width=0.48\textwidth]{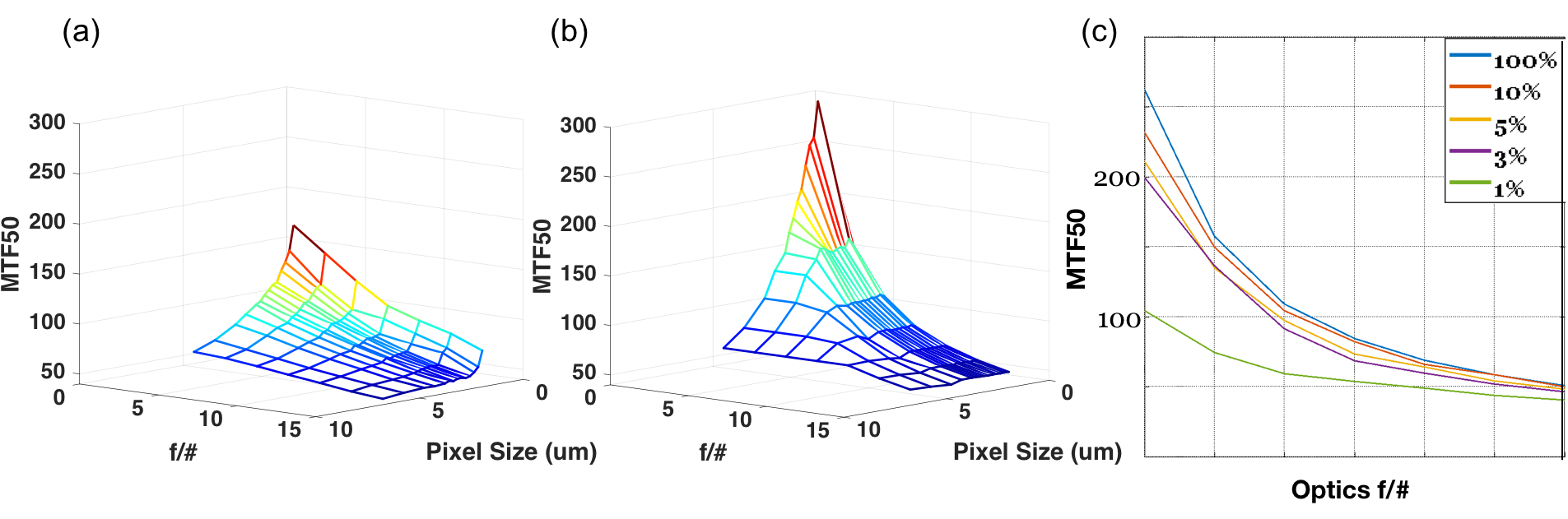}
\end{center}
\caption{\textbf{Dependence of spatial resolution (MTF50) on system and acquisition conditions.} (a) The MTF50 depends on both the f/\# of the diffraction-limited optics and the pixel size. The mesh shows the MTF50 for a range of these parameters. The transforms are learned from images  of human face images. (b) MTF50 of the transform as function of f/\# and pixel size when trained with a spatial resolution test chart. (c) MTF50 as a function of f/\# plotted separately for classes at different percentages of the maximum response level.}
\label{Fig:L3MTF50}
\end{figure}

We assessed the spatial resolution for different lens (diffraction-limited) and sensor combinations (Figure \ref{Fig:L3MTF50}a). The radius of the blur circle grows proportionally with the f/\#, blurring the optical irradiance at the sensor. When the pixel size is small and the f/\# is large, the spatial resolution, assessed by MTF50, is limited by the optics. When the pixel size is large and the f/\# is small, the spatial resolution is limited by the pixel size.

The exact value of the MTF50 also depends on the training data. In Figure 10a, the $L^3$ transforms are trained with a collection of human faces which do not have sharp edges. To evaluate the upper limit of spatial resolution, we trained with a scene containing only a spatial resolution chart and again used the MTF50 metric to evaluate spatial resolution (Figure \ref{Fig:L3MTF50}b). In this case, the $L^3$ transforms achieve almost double the spatial resolution in the optimal region (small pixel size, small f/\#). In other regions, the benefits of using spatial resolution target is much smaller.

In addition to the optics and pixel size, scene illumination level and exposure conditions also matter (Figure \ref{Fig:L3MTF50}c). We evaluated the MTF50 for a fixed pixel size using over a range of response levels and f/\#s. For small f/\# when the resolution can be very high, the $L^3$ transforms differ between the response levels. At low response levels, the transforms reduce noise by placing significant weights on most of the pixels in the patch. Thus the MTF50 is relatively low. When the response level is high, the MTF50 is much higher. For large f/\# the loss is dominated by the lens and thus the difference in the MTF50 between low and high response levels is minimal.

\subsubsection{RGB-NIR}
Next, we use the $L^3$ method to design an image processing pipeline for an RGB near infrared (NIR) sensor. The main application for including an NIR channel is to acquire extra information that is used in combination with an IR projector to estimate depth \cite{Jeong13, Smisek13}. There are several ways to implement NIR sensors. One approach removes one of the two green filters and the IR cutoff filter that is normally placed on the sensor surface. Modern color filters pass significant amounts of IR, so this approach allows NIR photons to enter the same pixels that are used by the visible light\cite{Fredembach13}. The image processing pipeline must estimate and remove these correlated IR signals, which introduces noise and reduces sensor dynamic range.

An alternative approach, recently implemented by Panasonic, selectively blocks IR photons in the RGB channels placing metal within the pixel\cite{Watanabe15}. In this approach, each CFA block contains a pixel of each type, and the RGB channels are protected from absorbing NIR photons by an infrared cut filter layer which includes a stack of silicon oxide and titanium oxide films.

The Panasonic RGB-NIR differs significantly from RGBW because the NIR channel captures very little information in the visible range. However, there may be useful image reproduction information in the NIR channel, and in any case the pipeline must run effectively even if the imaging component only uses the RGB channels. 

\begin{figure}[!htbp]
\begin{center}
\includegraphics[width=0.48\textwidth]{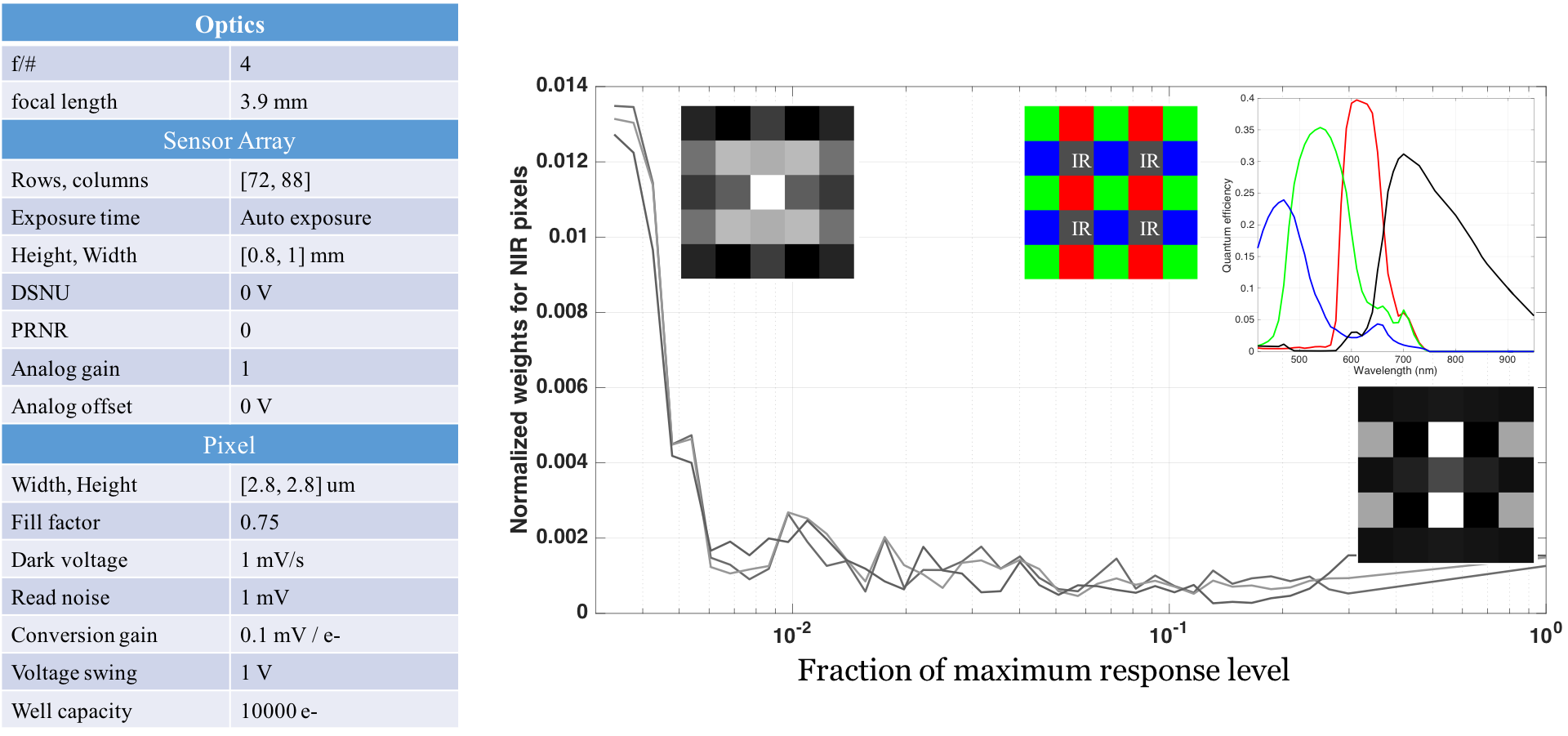}
\end{center}
\caption{\textbf{Key parameters of RGB-NIR camera systems (left) and the normalized weights for NIR pixels as a function of response levels (right).} The three curves show the normalized weights for NIR pixels towards three output channels as a function of response levels. The two inset gray images are the learned transforms for a G centered patch towards blue output under low and high response range. In low light, the NIR pixels are slighted used while in high response levels, there is almost no contribution from NIR pixels in the RGB rendered image. The CFA pattern and effective quanta efficiency of the pixels (CFA transmittance included) are shown as the inset image on the upper right.}
\label{Fig:L3RGBNIR}
\end{figure}

In this example, we simulate the Panasonic design and the key parameters are shown in Figure \ref{Fig:L3RGBNIR}a. Learning the $L^3$ pipeline for this sensor requires a hyperspectral scene radiance dataset that extends into the NIR. We used a dataset of 12 scenes that were measured from 415 to 950 nm. The data set includes calibration targets, fruits, buildings and natural scenes.

We used the $L^3$ method to learn a set of local linear transforms for the RGB-NIR sensor. Figure \ref{Fig:L3RGBNIR} shows the normalized weights of NIR pixels for G centered patches at different response levels. For low response levels, the $L^3$ transforms assign significant weight on the NIR pixels; the weights on these pixels become very small at high response levels. 

We evaluated the color accuracy for the RGB-NIR sensor as we did for the RGBW case. The cross-validated median CIELAB $\Delta E$ value is 2.87. If we black out the IR pixel and solve for the transforms again, the cross-validated median $\Delta E$ increases slightly to 3.08. This shows that the NIR data can slightly help estimate color. If we replace the IR sensor with a green pixel, forming a conventional Bayer pattern, the median $\Delta E$ value is 2.69. Hence, the RGB-NIR acquires some information in the invisible range at a cost of slightly worse RGB image. The simulations and $L^3$ algorithm quantify how to take advantage of this information in natural indoor images.

We also evaluated the spatial resolution of the RGB-NIR design using the MTF50 measure. For the Panasonic camera, the MTF50 value is 137 cycles/mm (optics f/\#=4, 2.75 um pixel size with proper exposure). Replacing the NIR pixel with a G pixel, increases the MTF50 to 151 cycles/mm. This shows that the RGB-NIR design has slightly lower spatial resolution than the matched Bayer design.

\section{Discussion}
We quantified how accurately the $L^3$ architecture renders images. We used a perceptual error metric to compare the $L^3$ rendering with two different commercial rendering methods (cf. Nikon and DxO). The $L^3$ architecture, based on kernel regression approximation, produces images that are about 2 $\Delta E$ (spatial CIELAB) from the original.

We also showed that the $L^3$ architecture can automatically generate rendering pipelines that are optimized for novel sensor designs. We implemented and evaluated image rendering pipelines for a sensor designed to extend the dynamic range by including a clear pixel (RGB-W), and also for a sensor that includes a set of pixels capable of measuring near infrared patterns projected into the scene to estimate depth (RGB-NIR). 

In this section, we review the relationship between $L^3$ and other new ideas in image processing. Then, we discuss some design choices of the $L^3$ method that arise in practice. First, we discuss the need to create multiple tables of transforms that should be applied in different acquisition conditions (color balancing; exposure duration; imaging conditions). Second, we describe how we account for knowledge about the transformations that should be incorporated to improve the learned transforms. Third, we consider the choice of a target space for the rendering. Fourth, we discuss how $L^3$ might be extended to address additional modules in the rendering pipeline.

\subsection{Related work}
There are several themes in the literature that share common elements with the $L^3$ method. At the most general level, Milanfar et al. proposed using the multidimensional kernel regression and non-parametric learning methods in image processing and reconstruction \cite{Takeda07, Milanfar11}. That work generalizes several image processing methods, including bilateral filtering, and denoising algorithms, under kernel regression schema. The general principles of kernel regression - classifying local data and interpolating measurements - can be applied to a range of imaging problems, such as learning super-resolution kernels \cite{Zhang15}. 

The proposal that is closest to our work comes from Khabashi et al. \cite{Khashabi14} Similar to our work, they describe a nonparametric regression tree models, together with Gaussian conditional random fields, to demosaic raw sensor response. They classify the data near each pixel into one of a large number of classes; the class is based on the color filter type of the pixel and a measurement of the local edge direction in the $5\times 5$ neighborhood of the pixel. For each class they use example camera data to find a quadratic transform that maps pixel data to the rendered value. 

In addition to these similarities in the approach, there are several significant differences. First, Khabashi et al. perform their training by processing mosaicked sensor data from existing cameras to estimate the ground truth, full resolution. In contrast, $L^3$ makes extensive use of image systems simulation technology to create sensor data for training. The use of simulations enables $L^3$ to support analyses for cameras that do not yet exist. Second, Khabashi et al. classify the sensor data based on pixel type and spatial orientation of the data in the of $5\times5$ patches within the sensor mosaic. $L^3$ classifies the sensor data based on response level and local contrast. Relying on response level is important because the different levels have very different noise characteristics, and the optimal transform differs significantly between low and high response levels (Figure \ref{Fig:L3TransformRGBW}). Finally, Khabashi et al. focus on the demosaicking stage of the process, while $L^3$ training replaces additional pipeline components, including denoising and color transforms that map the sensor data directly into the target color space.

Another related set of ideas concerns the development of image processing pipelines that are based on joint optimization across optics, sensor, and display. An example is from Stork and Robinson \cite{Stork08} who offered a theoretical foundation for jointly optimizing the design and analysis of the optics, detector, and digital image processing for imaging systems. They optimized the image processing pipeline for different lenses, assuming a monochrome sensor. The $L^3$ method incorporates lens properties into the simulation, so that the table of transforms accounts for the specific lens properties. Different tables are generated as the lens properties (e.g., aperture, f/\#) are varied. Hence, the $L^3$ method is also a co-design approach in the sense that the learned rendering parameters depend on the whole system, including the optics and sensor.

Heide et al. also conceive of the image processing pipeline as a single, integrated computation\cite{Heide14}. They suggest a framework (FlexISP) in which they model the relationship between the sensor data and a latent image that represents the fully sampled sensor data prior to optical defocus. They propose to estimate the latent image from the sensor data by solving an optimization problem; the optimization accounts for both the data and a set of natural image priors. Hence, a key difference is that Heide et al. calculate a solution separately for each sensor acquisition, while $L^3$ pre-computes a fixed table of transforms and applies this table to all images. Another difference is that Heide et al., like Khabashi et al., begin their calculations with the sensor data. In contrast, $L^3$ simulates a camera system beginning with scene radiance, accounting for properties of the optics, pixel, and sensor. By working from scene spectral radiance data, $L^3$ can be used to create pipelines at the earliest stages of the design process, when no hardware implementation yet exists. The simulations also make it possible to optimize $L^3$ parameters for different types of scenes, some of which may be difficult to create in a laboratory environment.

Convolutional sparse coding (CSC) methods share some features of the $L^3$ method. CSC representations begin with a full image representation and decompose the image into a linear sum of component images \cite{Heide15}. Each component is the convolution of a single, usually small, kernel with a sparse feature map (most entries are zero). The CSC learns local features from the input training images, and the core calculations are linear. However, the CSC learning methods and target applications of the differ significantly from $L^3$. First, CSC learns kernels and feature maps that decompose an image into separate components. $L^3$ performs the reverse computation; it starts with partial sensor data and creates a complete image. Second, the learning methods are different. The CSC kernels are learned through advanced bi-convex optimization methods that require substantial computational power. The affine (or simple polynomial) transforms learned by $L^3$ use prior knowledge of the camera and training about the camera design but very simple optimization methods. In summary, $L^3$ is an architecture for designing new image processing pipelines and efficient rendering; CSC is a technique for feature extraction and applications to learning image features for machine vision applications and computational photography, such as inpainting.

\subsection{Multiple transform tables}
Training $L^3$ for the range of settings (optics, sensor properties) of a single camera, leads to different transform tables (Figure 10). For mobile devices, the camera settings do not vary extensively. When only a few number of possible settings are available, the best solution might be to pre-compute and store a table of transforms for each setting.

In addition to hardware settings, there is the question of whether the $L^3$ training would produce different tables as we change the scene characteristics. We have explored important case: how the tables depend on the spectral power distribution of the illumination {Germain}. In this case we found that the tables learned for different illuminant are similar enough so that we can render the image with a single table and then apply a $3\times3$ color transform to render a color-balanced image.

There are many other different scene characteristics that remain to be explored. The optimal table of transforms may depend on factors such as image motion, image content, optics parameters such as depth of field and focus. It is possible that multiple tables will be required or that a single set of tables followed by simple transforms will suffice for most conditions.

\subsection{Applying prior knowledge to the transform table}
$L^3$ training depends on the specific training data. This can be used to our advantage, say if we know we want to optimize the rendering for a particular condition and target (e.g., human faces, outdoors). The reliance on specific samples produces transforms that may differ in some small way from our expectations. For example, in many cases we expect the transforms to be left-right symmetric. Further, we expect that transforms at nearby response levels will be similar to one another. We developed functions that can be applied to the learned table of transforms to enforce these expectations (prior knowledge).

\begin{itemize}

\item \textbf{Symmetry}

When the underlying color filter array of each patch is symmetric in some way (up-down, left-right, transpose, circular), we also expect the learned transform to be symmetric. Imposing symmetry helps avoid over-fitting to the training data. We transform general transforms into symmetric transforms by creating symmetric versions of the learned transform and then using the average.

\item \textbf{Smoothing and interpolation}

The $L^3$ coefficients change relatively smoothly as the response level increases. We smooth the transforms by fitting a spline to each of the coefficients and then replacing the coefficients with the value of the smooth spline. We use the same method to interpolate for transforms in classes that have a small amount or insufficient training data.

\item \textbf{Uniformity}

A uniform scene should be rendered as a uniform image. This requirement, unlike the previous two, requires operating on the transforms from different pixel types. Specifically, the sum of the transform weights of each pixel type must be equated between the classes of different center pixels.

\end{itemize}

\subsection{Choosing the target space for rendering}
We emphasized $L^3$ consumer photography applications: the sensor data are transformed to a rendered image. Even for consumer photography applications, there are multiple choices for the target rendering space. We have trained $L^3$ instances to transform into various colorimetric spaces (e.g., CIE-XYZ), and we have also trained to transform into nonlinear representations (e.g., sRGB, CIELAB). Because the global $L^3$ transformation is nonlinear (though locally linear), the input data can be effectively transformed to most representations that are smoothly related to colorimetry representations. Choosing the target space is equivalent to choosing the error function. For example, rendering to CIELAB space minimizes the point-by-point color error. 

\subsection{Space-varying}
The current $L^3$ formulation does not account for the position of the center pixel within the sensor. Thus, the algorithm is effectively shift-invariant. There are two important aspects of the rendering pipeline that are space-varying. First, lens shading produces an uneven illumination level from the center to the periphery of the sensor. Second, geometric distortion of the image (e.g., barrell distortion) varies the relationship between the position of the pixel within the sensor and its appropriate position in the output image. 

The pixel type and response level are used to identify a class, and we do not include the position of the pixel within the sensor. Hence, the $L^3$ method is fundamentally space-invariant. Correcting for the shift-varying components (lens and geometric distortion) are shift-varying. The parameters of these features are determined by the main taking lens and are independent of the image processing pipeline. Hence, like illuminant correction, for the moment we think it is best to perform these steps separately rather than extending the number of classes and setting $L^3$ the task of accounting for the position-dependent factors.

\subsection{Reproducibility}
The data and methods necessary to reproduce the figures are available from the Stanford Digital Repository\footnote{http://purl.stanford.edu/bk962py0458}.

\section{Conclusion}
We introduce a methodology to automate the design of image processing pipelines. The image-processing pipeline is approximated as a locally linear operation in which sensor data are grouped into various classes, and the data from a class are rendered by a linear transform into the rendered image. We illustrate that the local transforms can produce high quality rendered images. Then, we use image systems simulation to create the table of affine transforms for novel camera designs, including sensors with clear or near infrared pixels. We evaluate the performance of these tables using color metrics (S-CIELAB), spatial resolution metrics (MTF50), and simulations of captured images. Hence, this paper combines image systems simulation technology and modern computational methods into a methodology that creates image processing pipelines.



%

%


\end{document}